\documentclass[12pt]{article}
\usepackage{authblk}
\usepackage{eucal}
\usepackage{mathrsfs}
\usepackage{amsmath}
\usepackage{amsfonts}
\usepackage{amssymb}
\usepackage{ifluatex}
\ifluatex
\usepackage{fontspec}
\defaultfontfeatures{Ligatures=TeX}

\usepackage[]{unicode-math}
\unimathsetup{math-style=TeX}
\else 
\usepackage[utf8]{inputenc}
\fi 
\ifluatex\else\usepackage{stmaryrd}\fi
\usepackage{url,multirow,morefloats,floatflt,cancel,tfrupee}
\usepackage{textcomp}
\usepackage{booktabs}
\usepackage{graphicx}
\usepackage[version=4]{mhchem}
\usepackage{siunitx}
\usepackage{longtable,tabularx}
\usepackage{hyperref}
\usepackage{diagbox}
\usepackage[noadjust, compress]{cite}
\usepackage[linesnumbered,ruled,vlined]{algorithm2e}
\SetKwInput{KwInput}{Input}                % Set the Input
\SetKwInput{KwOutput}{Output}              % set the Output

\title{Graph Convolutional Neural Networks for Body Force Prediction}

\author[1]{\normalsize Francis Ogoke}
\author[1]{Kazem Meidani}
\author[2]{Amirreza Hashemi}
\author[1,3,4,$^\ast$]{Amir Barati Farimani}
\affil[1]{\footnotesize Department of Mechanical Engineering, 
    Carnegie Mellon University, Pittsburgh, PA, USA}
\affil[2]{\footnotesize Department of Computational Modeling and Simulation, University of Pittsburgh, Pittsburgh, PA, USA}
\affil[3]{\footnotesize Department of Chemical Engineering, 
    Carnegie Mellon University, Pittsburgh, PA, USA}
\affil[4]{\footnotesize Machine Learning Department, 
Carnegie Mellon University, Pittsburgh, PA, USA}
\affil[$^\ast$]{\footnotesize Corresponding author, E-mail:  barati@cmu.edu.}

\date{}

\begin{document}

\maketitle

\begin{abstract}
Many scientific and engineering processes produce spatially unstructured data. However, most data-driven models require a feature matrix that enforces both a set number and order of features for each sample. They thus cannot be easily constructed for an unstructured dataset. Therefore, a graph based data-driven model to perform inference on fields defined on an unstructured mesh, using a Graph Convolutional Neural Network (GCNN) is presented. The ability of the method to predict global properties from spatially irregular measurements with high accuracy is demonstrated by predicting the drag force associated with laminar flow around airfoils from scattered velocity measurements. The network can infer from field samples at different resolutions, and is invariant to the order in which the measurements within each sample are presented.  The  GCNN method, using inductive convolutional layers and adaptive pooling, is able to predict this quantity with a validation $R^{2}$  above 0.98, and a Normalized Mean Squared Error below 0.01, without relying on spatial structure.
\end{abstract}

\section{Introduction}
Due to recent advances in data-driven methods and the proliferation of scientific data, there has been a significant amount of attention towards data-driven inference to model or predict system properties. This is particularly relevant in fluid mechanics, where large amounts of data are needed to understand potentially complex, multiscale flow phenomena. The success of Deep Learning (DL) in computer vision has inspired its application in studying physical phenomena. Physics informed Neural Networks are used to learn the physics behind and solutions to high-dimensional Partial Differential Equations (PDEs) \cite{Han8505,Lye_2020} \cite{RaissiPINN_FR,pang2020npinns,JAGTAP2020CPINN}. Deep Neural Networks (DNNs) have been of particularly high interest in surrogate modeling and predicting complex transport phenomena \cite{sharma2018weaklysupervised,kim2020prediction}. Wiewel et al. used a latent space learning to efficiently simulate the temporal evolution of the pressure field \cite{wiewel2018latentspace}. Farimani et al. applied conditional Generative Adversarial Networks (cGAN) to solve the physics of transport phenomena without knowledge of the governing equations \cite{farimani2017deep}.

Machine learning has seen success in generating flow fields based on data collected from experiments, and noisy data from numerical simulations \cite{kissas2019machine, maulik2020probabilistic, YAN2019115, cai2019dense}. For example, Particle Image Velocimetry (PIV), where velocity fields are generated by tracking the movement of tracer particles, has been introduced as a non-intrusive technique for analyzing flow behavior and measuring the forces interacting with an immersed object \cite{noca1999comparison,graham2017impulse}. The analysis of PIV data using Machine Learning has allowed for more efficient flow field reconstruction and prediction. Rabault et al. used a Convolutional Neural Network (CNN) architecture to surrogate PIV by cross-correlating point locations between two frames, therefore predicting the flow velocity field \cite{rabault2017performing}. Morimoto et al. applied a CNN to artificial PIV data to develop a method for reconstructing flow field from snapshots with missing regions \cite{morimoto2020experimental}.

Various machine learning algorithms have recently been applied to structured flow field data to facilitate predictions based on immersed flows around streamlined objects, such as airfoils. For example, the leading-edge suction parameter (LESP) and angle of attack (AoA) are of high importance for discrete vortex methods to be effective. Hou et al. used a combination of convolutional neural networks and recurrent neural networks to predict these parameters based on time-dependent surface pressure measurements \cite{hou2019machine}. In related work, Provost et al. used the same method to optimize the number of sensors necessary for LESP and AoA prediction \cite{le2020deep}.

Zhang et al. applied convolutional neural networks on image representations of various airfoils and their surrounding flow to predict the lift coefficient. The airfoils are immersed in different flow conditions, and the parameters of the flow (e.g. Mach Number) are encoded as pixel intensities \cite{zhang2018application}. Viquerat and Hachem used an optimized CNN to estimate the drag coefficient of several arbitrary 2D geometries in laminar flow. A large training sample of random 2D shapes along with their drag forces computed by immersed mesh method were used to increase the prediction accuracy of realistic geometries such as NACA airfoils \cite{viquerat2019supervised}. Yilmaz and German used a CNN to predict airfoil performance directly from the geometry of the airfoil, replacing cumbersome surrogate methods that required manual parameterizations of the airfoil using shape functions \cite{yilmaz2017convolutional}.
%on airfoil geometry for prediction of airfoil performance to overcome the drawback of surrogate modeling methods which is the requirement for shape parametrization.

Guo et al. trained a deep CNN to make fast but less accurate visual approximations of the steady state flow around 2D objects which improves the design process by expediting the alternatives generation \cite{Guo}. Miyanawala and Jaiman used a CNN to predict aerodynamic coefficients for several bluff body shapes at low Reynolds numbers. They used structured data of an encoded distance function to predict unsteady fluid forces \cite{miyana}. Bhatnagar et al. also used a signed distance function as well as a limited range of both Reynolds numbers and angles of attack to predict flow field velocities and pressure for several airfoils \cite{bhatnagar2019prediction}.

These methods, however, are limited in their ability to generalize to unstructured data. As traditional machine learning methods require the creation of a feature matrix with both a specific size and order of input samples, they cannot be applied to unstructured data. Flow field data, however, can be highly unstructured due to the use of irregular meshes to define curved or complex geometries.

Recent interest in manipulating unstructured data has led to the development of both mesh-free inference methods for point cloud representations and reduced-order models based on graph-based representations of fluid data \cite{gross2020meshfree}. Several works have used graph theory-based methods to identify coherent structures within turbulent flow \cite{schlueter2017coherent, padberg2017network}. Hadjighasem et al. examined the generalized eigenvalue problem of the graph Laplacian to develop a heuristic for determining the locations of coherent structures \cite{hadjighasem2016spectral}. This work is extended to extract coherent structures from the vortical behavior of the flow by Meena et. al, where a graph is constructed to represent the mutual interaction of individual vortex elements, and larger vortex communities are identified with network theory-based community detection algorithms \cite{meena2018network}.  
To perform mesh-free inference, Trask et al. introduce the idea of GMLS-Nets. GMLS-Nets parameterize the generalized moving least-squares functional regression technique for application on mesh-free, unstructured data. Trask abstracts the GMLS operator to perform convolution on point clouds. They demonstrated success in both uncovering operators governing the dynamics of PDEs and predicting body forces associated with flow around a cylinder based on point measurements \cite{trask2019gmls}.

In this paper, we present a method for data-driven prediction from flow fields defined on irregular and unstructured meshes, using a Graph Convolutional Neural Network (GCNN) framework. GCNNs have been applied to problems dealing with unordered data points where specific relationships between the points encode important information and thus are often used in applications such as natural language processing, traffic forecasting, and material property prediction\cite{battaglia2018relational, bastings2017graph, monti2016geometric, Yu_traffic2018, yao2019graph, PhysRevLett.120.145301, chen2019graph, duvenaud2015convolutional}. Graph Neural Networks have also previously been used to model spatiotemporal phenomena and surrogate physics simulators \cite{sanchezgonzalez2020learning, alet2019graph}. Our approach exploits the irregular mesh that the velocity field is defined on to regress from unordered data across both varying resolutions and non-uniform spatial distributions.

In section 2, we first present our methodology of using graph representation of the unstructured mesh. The learning algorithm based on an inductive graph convolution method is then discussed and finally, the structure of the implemented network is explained. In section 3, The data which is used in our study and the results for the corresponding experiments are discussed and compared to other traditional methods. A conclusion of the work and suggestions for some possible future directions are also provided in section 4.

\begin{figure*}[!htbp]
\begin{center}
   \includegraphics[width=1\linewidth]{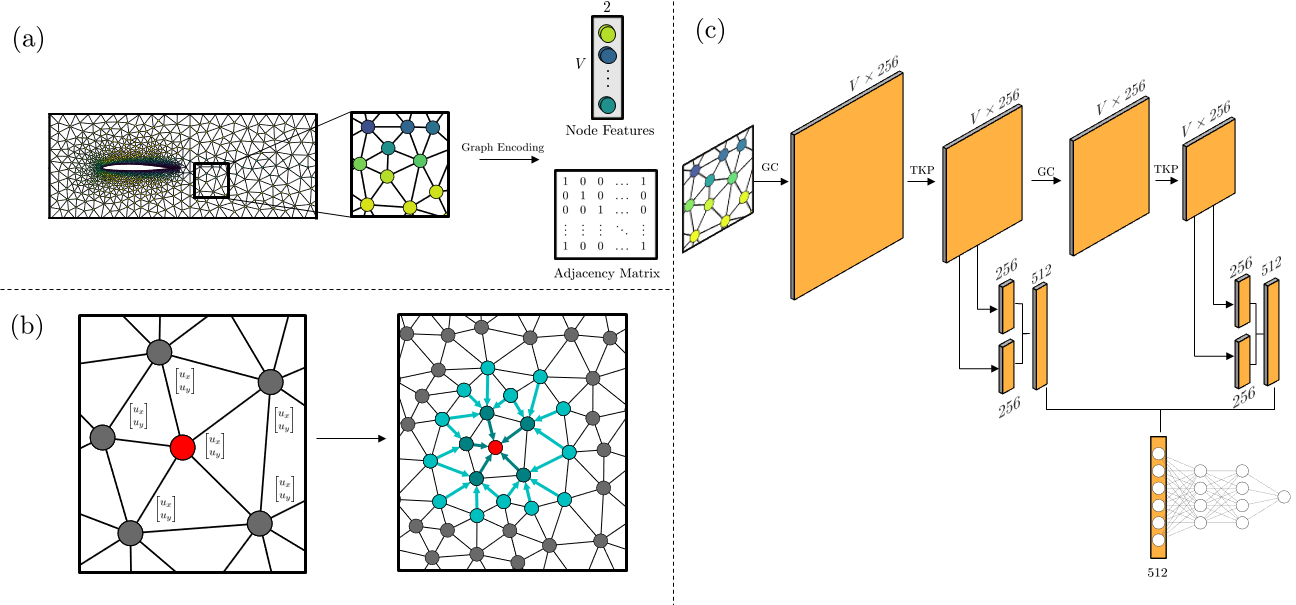}
\end{center}
   \caption{\textbf{a)} The velocity information defined on the unstructured mesh is represented as a graph, where the mesh nodes are vertices of the graph, and the connectivity of the mesh is taken as the edges of the graph. An $2 X N$ matrix of node features contains the velocity in each dimension, and an $N X N $ adjacency matrix encodes the connectivity of the matrix. 
   \textbf{b)} The Graph Convolution operation. (left) The graph before a convolution operation is performed on the center node (red). (right) During graph convolution, the information in each of the rings of \textit{N}-order neighbors, where $N \leq k$, is aggregated to the center node. In this application, $k = 2$. 
   \textbf{c)} The architecture of the Graph Convolutional Neural Network. 'GC' refers to the graph convolution operation in b), 'TKP' refers to Top-K Pooling. The feature map output of each top-K pooling layer undergoes both mean pooling and max pooling, and the outputs of each operation are concatenated together. The concatenated output from each layer is added together and passed to fully connected layers for regression.}
\label{fig1}
\end{figure*}

\section{Methods}
\subsection{Graph Representation}

% The key idea behind our approach is to use a graph representation to describe the connectivity of the data points, without relying on the nodes of the graph having a structured representation that is consistent between samples. This approach is necessary, as without a predefined ordering of the nodes, there is no way to construct a traditional feature matrix with consistent values between each sample. Therefore, a graph representation is necessary to represent the unstructured grid.

% We applied various machine learning algorithms on the data both with and without a graph featurization, and our results show the importance of the graph featurization. The nodes in each airfoil mesh have different spatial positions, so raw features (i.e. node velocities) are not defined in the same way between sample points. However, the mesh generation process around the airfoils follows a specific order, which allows us to create a feature matrix. Following the predetermined order of node ID's from GMSH places each node at similar positions between the different samples, making that makes training on the raw data possible with shallow machine learning algorithms. But if the order of the nodes are shuffled, predictably, no learning will be observed with these algorithms. 

The key idea behind our approach is to use a graph representation to describe the connectivity of unstructured data points. In order to resolve flow fields around complex geometries in computational fluid dynamics,  an irregular mesh around the immersed object is created. Next, numerical methods are applied to calculate the flow field data. Any mesh structure around the immersed object in the flow field can be represented as a graph, considering mesh nodes as vertices and using edges to connect the neighbors. Therefore, one can construct a graph representation of the unstructured mesh with different complexities or number of nodes around arbitrary objects. 

 Specifically, we define an undirected graph $\mathcal{G} = (\textbf{V}, \textbf{E})$, with $\textbf{V}$ vertices describing the nodes of the mesh, and $\textbf{E}$ edges representing the connectivity of the mesh. The flow field data is defined on mesh nodes, resulting in a feature matrix that contains the input features for \textit{V } graph vertices. The edge connections are encoded as the adjacency matrix, a binary $V\times V$ matrix indicating whether any given pair of vertices are connected. 

%whether a pair of nodes in the graph are adjacent or not. In our application, as meshes have repeating patterns of local node adjacency, this encoded matrix is highly sparse. (Figure 1.a) 

\subsection{Graph Convolution}

Graph Convolutional Neural Networks (GCNNs) are the generalization of Convolutional Neural Networks (CNNs) for operation on graphs. GCNNs, like  CNNs,  are able to extract multi-scale spatial features through the use of shared weights and localized filters \cite{kipf2016semi}. However, as discussed earlier, traditional CNNs are unable to work with unstructured data. GCNNs can bypass this limitation by defining the convolution operation based on the structure of the graph.  By propagating information through each node's local neighborhood as defined by the adjacency matrix, GCNNs are invariant towards the order in which the nodes are specified in the feature matrix. GCNNs are often used for tasks such as node classification, link prediction \cite{linkpred}, and graph classification. 

GCNNs can be described in terms of a general framework for learning on graph-structured data, called Message Passing Neural Networks (MPNNs) \cite{gilmer2017neural}.  MPNNs develop hidden state embeddings, $h_v$ for each node $v$ during the training process. Supervised training of a graph neural network aims to learn a state embedding from the features defined on the nodes and edges of the graph to have the best possible mapping to the output. The training process consists of two phases, the message passing phase where hidden states aggregate information from their surrounding nodes, and the readout phase where a feature vector for the graph is computed from the hidden states. The message passing process is parameterized by two functions, the message function $M_t$ and the node update function $U_t$, while the readout function is given by $R$. $R$, $M_t$, and $U_t$ are all learned differentiable functions that are updated during the training process.  

Defining $e_{vw}$ as the edge connecting node $v$ to node $w$ and $\mathcal{N}(v)$ as the neighborhood of node $v$, the message passing phase can be formalized as:

\begin{equation}
\begin{aligned}
  m_v^{t+1} = \sum_{w \in \mathcal{N}(v)} M_t(h_v^t, h_w^t, e_{vw})
  \\
  h_v^{t+1} = U_t(h_v^t, m_v^{t+1})
\end{aligned}
\end{equation}

Based on $h_v$, the readout phase computes a feature vector as:

\begin{equation}
\begin{aligned}
\hat{y} = R({h_v^T | v \in G})
\end{aligned}
\end{equation}

A standard framework used is the Laplacian based GCNN, detailed in \cite{kipf2016semi}, where $M_t$ and $U_t$  are defined as the following: 

\begin{equation}
\begin{aligned}
M_t(h_v^t, h_w^t) = \tilde{A}_{vw}(deg(v)deg(w))^{-\frac{1}{2}}h_{w}^t
\end{aligned}
\end{equation}
\begin{equation}
\begin{aligned}
U_t(h_v^t, m_v^{t+1}) = \sigma((W^{t})^{T}m^{(t+1)})
\end{aligned}
\end{equation}

where $\tilde{A}_{vw}$ is the adjacency matrix describing the connectivity of the graph, assuming that each node is connected to itself, and $deg(v)$ is the number of nodes connected to node $v$ \cite{gilmer2017neural}.

The GraphSAGE method, introduced by \cite{hamilton2017inductive} is in close relation to the Laplacian based GCN layers described in the message passing formalization above. In the version of GraphSAGE implemented in this paper, the GraphSAGE algorithm is the inductive variant of this message passing network, with specific modifications to improve the accuracy and efficiency of the model. GraphSAGE acts in an inductive manner, operating on each node rather than on the entire graph, as it uses each node's local neighborhood in order to learn a function that can generate appropriate node embeddings.  This method first samples a fixed number of nodes from the $k$ nearest neighbors of each node and then applies an aggregation operator to transfer information to the node itself (Algorithm 1). The aggregator can be a weighted averaging operation with trainable parameters. This inductive framework is especially helpful in the case of large graphs where low-dimensional embeddings of the nodes are more important \cite{hamilton2017inductive}. In this regression application, the readout phase consists of a fully connected neural network that predicts a single global value for each graph based on the hidden state embeddings.

\begin{algorithm}[H]
\DontPrintSemicolon
  
  \KwInput{Graph $\mathcal{G} = (\textbf{V}, \textbf{E})$; input features $\{x_{v},\forall{v}\in \mathcal{V}\}$; depth $K$; weight matrices $\textbf{W}^k,\forall{k}\in \{1,...,K\}$; non-linearity $\sigma$; neighborhood function $\mathcal{N}(v) = \{u\in   \mathcal{V} : (u , v) \in \mathcal{E}\} $} 
  %differentiable aggregator functions ${\scriptstyle AGGREGATE_k} , \forall{k}\in \{1,...,K \}$
  \KwOutput{Vector representations $z_{v}$ for all $v\in \mathcal{V}$}
%   \KwData{Testing set $x$}
  $h_{v}^{0} \gets x_{v},\forall{v}\in \mathcal{V}$ \\
  
  \For{$k=1...K$}
    {
        \For{$v\in \mathcal{V}$}
        {

        $h_{v}^{k} \gets \sigma(\textbf{W}^{k}\cdot{\scriptstyle MEAN}(\{h_{v}^{k-1}\} \cup \{ h_{u}^{k-1},\forall{u} \in \mathcal{N}(v)\}))$ %h_{\mathcal{N}(v)}^{k}));$
        % \EndFor
        }
        $h_{v}^{k} \gets h_{v}^{k} / \|{h_{v}^{k}}\|_{2}, \forall{v} \in \mathcal{V}$
    % \EndFor
    }
  $z_{v} \gets h_{v}^{K},\forall{v}\in \mathcal{V}$ 

\caption{GraphSAGE embedding generation algorithm, reproduced from \cite{hamilton2017inductive}}
\end{algorithm}

Flow field meshing usually results in a relatively large graph around the objects in comparison to other common applications such as molecular graphs. Hence, we use a node level embedding graph convolution operator that is based on the average aggregator in the GraphSAGE framework. Since the mesh has a specific edge connection pattern in which each node is only connected to a few neighbors, the sampling operator is not used. In our convolution operation, the features in the $k$ nearest rings in the neighborhood of each node are transferred to the center node by a trainable aggregation operation (Figure 1b, line 4 in Algorithm 1). Here, we use the Pytorch Geometric \cite{fey2019fast} library to load the data and implement the graph convolutional layers and pooling. 

\subsection{Network Architecture}
For this problem, we implement a Graph Convolutional Neural Network using GraphSAGE convolutional layers and top-K pooling steps, similar to the approach described in \cite{cangea2018towards}. Different from \cite{cangea2018towards}, we use an inductive convolution as opposed to a transductive convolution. Inductive methods are capable of generalizing to graphs with different structures, here allowing for prediction on meshes with varying resolutions.  The inputs of the network are graphs with node level velocity features, while the output is the value of the predicted drag force. Specifically, we use two GraphSAGE layers, each followed with a top K pooling layer. Top K pooling is a downsampling method to reduce the size of the layers by selecting the most important features. In top K pooling layer, a learned score is assigned to each node, and the nodes with the $K$ highest scores are selected to be passed to the next layer \cite{gao2019graph}. The output of each pooling layer is pooled twice, once using global mean pooling and then using global max pooling, and the output from each pooling operation are concatenated together. While the input size to the network can vary in different samples as they have various number of nodes, the output size should be the same. By pooling along the dimension of vertices, the global pooling operations result in the same output size.  The pooled, concatenated vectors are added together as a "skip connection", to reinforce the information contained in the sparse convolved feature maps. The output from this step is then fed to a fully connected network with three hidden layers, that predicts the drag force (Figure 1c). The training details of the GCNN are provided in Table 1 and the parameters are defined in \cite{pytorch,fey2019fast} for interested readers.

\begin{table}[]
\begin{center}
\caption{GCNN Parameters}
\renewcommand{\arraystretch}{1.2}
\resizebox{\textwidth}{!}{\begin{tabular}{lll|lll}
\hline
Layers              & Parameters             & Settings  & Layers          & Parameters & Settings           \\
\hline
\multirow{3}{*}{Graph Convolution}  & Convolution width      & 64       & \multirow{3}{*}{Fully Connected} & depth      & 3                  \\
                    & depth (neighbor rings) & 2        &                 & width      & {[}256,128,64,1{]} \\
                    & Activation             & ReLU     &                 & Activation & ReLU               \\
\hline
\multirow{2}{*}{Top-K pooling}      & \multirow{2}{*}{ratio (k)}              & \multirow{2}{*}{0.5}      &   \multirow{2}{*}{Others}    &     Loss       &      MSE         \\
                    &             &      &                 & Optimiser & Adam              \\
\hline
\end{tabular}}
\end{center}
\end{table}

\section{Results}
\subsection{Data}

In order to test the performance of our method, we aim to predict the drag force on the airfoils directly from the unstructured flow field velocity. To generate the airfoils, coordinate files are extracted from the UIUC airfoil database which contains the cartesian coordinates outlining the shape of the airfoil \cite{cfd3a4869ab4439490668729fc1ba7ef}. The incomplete or non-meshable samples are then removed from the dataset. In addition, the geometries are normalized to have a unit chord length. Next, each airfoil coordinate file is imported to the open-source mesh generator GMSH \cite{geuzaine2009gmsh}. Meshes are created to reflect the variation in the density of information contained in the domain, with a finer mesh on the area close to the airfoil that resolves the complex boundary layer effects, and a coarser mesh further from the airfoil, where the flow is minimally affected by the presence of the airfoil. 

\begin{figure*}[!htbp]
\begin{center}
  \includegraphics[width=1.0\linewidth]{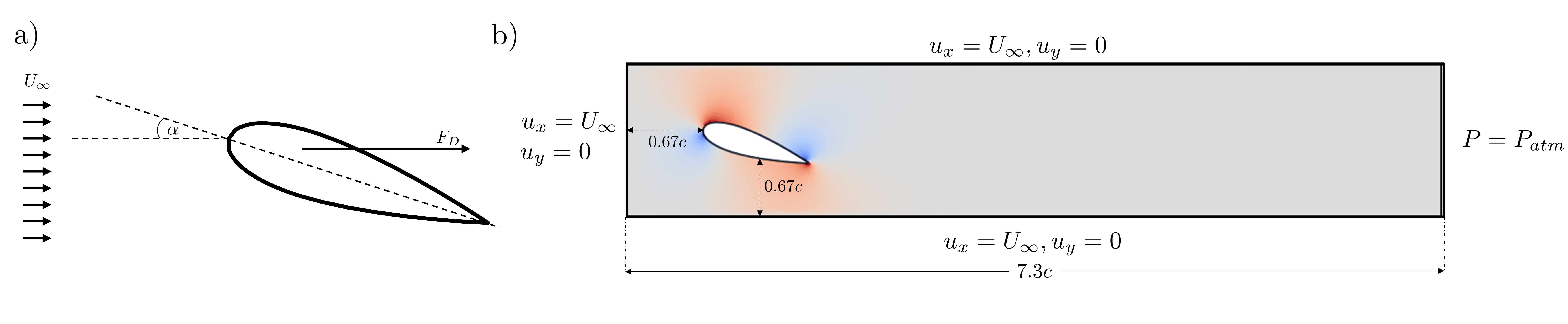}
\end{center}
  \caption{\textbf{a)} A sample airfoil from the dataset, at a \ang{10} angle of attack. The flow velocity past the airfoil is $U_\infty =$ 1.5 \si[per-mode=symbol]{\meter\per\second}
 \textbf{ b)} A schematic of the domain used to generate the data, where $c$ refers to the chord length of the airfoil. The airfoil is placed in a domain with a constant horizontal inflow velocity of $U_\infty =$ 1.5 \si[per-mode=symbol]{\meter\per\second}, and a pressure based boundary condition is used at the outlet of the domain. A free-slip boundary condition is used at the walls of the domain.  }
\label{fig2}
\end{figure*}

To compute the velocities at mesh nodes around each object and the corresponding drag force, we perform CFD simulations using the FEniCS \cite{Logg_2010} package. FEniCS supports the DOLFIN PDE solver, which is used to solve the incompressible Navier-Stokes equations with an Incremental Pressure Correction Scheme (IPCS) method \cite{thomadakis1996pressure}. The boundary conditions for these CFD simulations are a uniform velocity input of 1.5 \si[per-mode=symbol]{\meter\per\second} at the inlet (left), a far-field pressure condition at the outlet (right) and slip conditions at the top and bottom interfaces (Figure 2).The viscosity and the density of the flow are 0.001  $\mathrm{Pa}  \cdot \mathrm{s}$ 
and 1 \si{kg.m^{-3}} respectively. Selected airfoil samples with their velocity magnitude field along with their drag force are provided in Table 2. There is a positive correlation between the thickness of the airfoil and the corresponding drag force.

{ % begin box to localize effect of arraystretch change
\renewcommand{\arraystretch}{1.5}
\begin{table}
  [t] \caption{The velocity field and drag force for four different airfoil samples from the dataset} \centering \label{tab:stimuli}
  \begin{tabular}
      {lllll} 
      \hline \textbf{Airfoil Geometry} & \raisebox{-.5\height}{\includegraphics[width=1in]{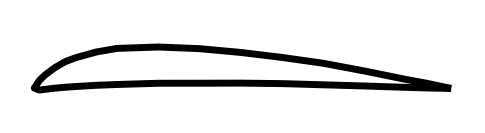}}  & 
      \raisebox{-.5\height}{\includegraphics[width=1in]{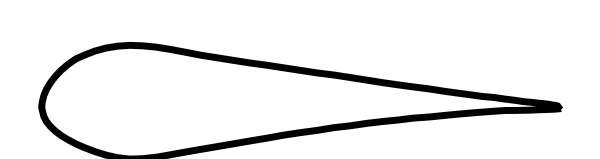}} &
      \raisebox{-.5\height}{\includegraphics[width=1in]{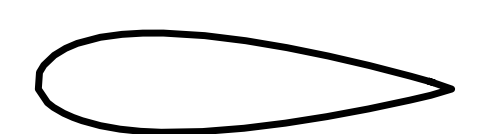}} &
      \raisebox{-.5\height}{\includegraphics[width=1in]{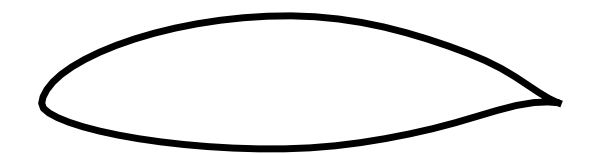}} \\ 
      %\addlinespace[0.5cm]
      \textbf{Velocity Field} & \raisebox{-.5\height}{\includegraphics[width=1in]{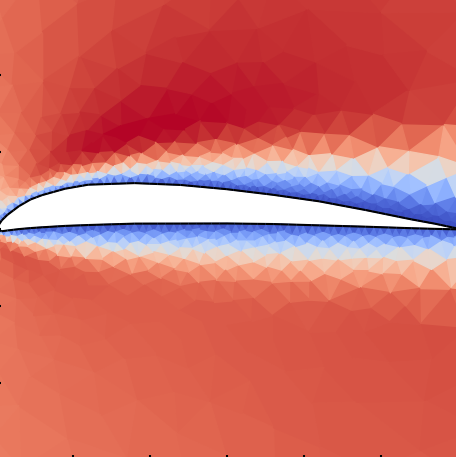}} & 
      \raisebox{-.5\height}{\includegraphics[width=1in]{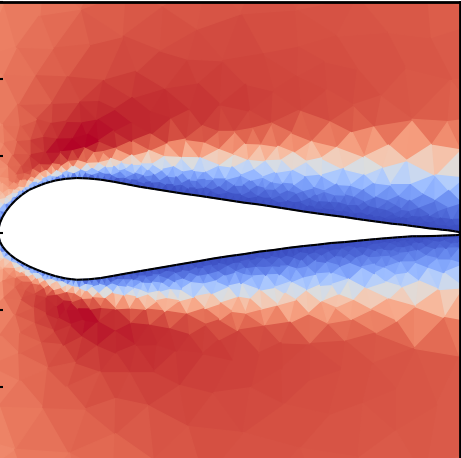}} &
      \raisebox{-.5\height}{\includegraphics[width=1in]{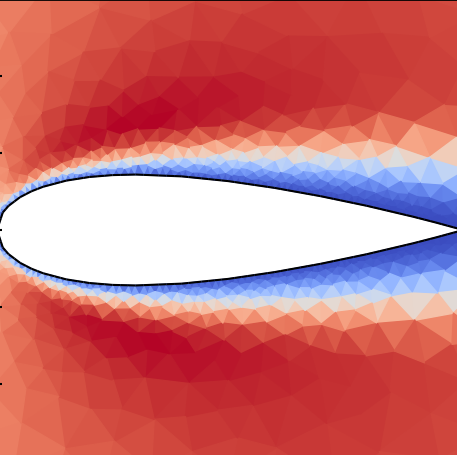}} &
      \raisebox{-.5\height}{\includegraphics[width=1in]{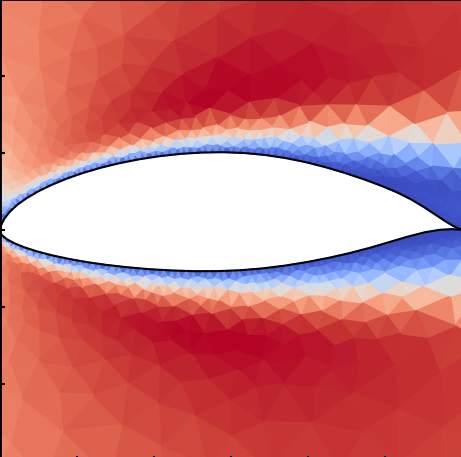}}   \\
      \textbf{$F_D$}& 
      $1.478$ & 
      $1.983$ &
      $2.274$ & 
      $2.462$ \\
      \hline
  \end{tabular}
%   \tabfnt{Note: Bold categories are biologically correct growth models.}
\end{table}
} % end box
 
The drag on an airfoil $A$ is calculated as follows:

\begin{equation}
 F_D = \int_A (\sigma \cdot n) \cdot e_x \mathrm{dS} 
\end{equation}

where $\sigma$ is the Cauchy stress tensor, $e_x$ is the horizontal unit vector, $n$ is the unit vector normal to the airfoil surface.

A grid convergence study is used to choose a specific mesh size to fully resolve the boundary layer effect while minimizing the computational time required. The resulted meshes contain 900-1500 mesh points and 3000-4000 edges, generated in effectively random spatial positions surrounding the airfoil.

% \subsection{Incremental Pressure Correction Scheme}
The incompressible Navier-Stokes equation are given by 
\begin{equation}
\rho (\frac{\partial \mathbf{u}}{\partial t} + u\cdot \nabla \mathbf{u} ) = \nabla \cdot \mathbf{\sigma}(\mathbf{u}, p) + f
\end{equation}
$$\nabla \cdot \mathbf{u }= 0$$

 To predict the velocity field at the next time-step ($u^{n+1}$)  from an existing time-step ($u^n$) while enforcing mass conservation, an \textit{Incremental Pressure Correction Scheme} (IPCS) is used to iteratively solve Equation 6. A detailed description of the IPCS method can be found in \cite{thomadakis1996pressure}.

The CFD outputs of each sample are then processed to generate the matrix of node level horizontal and vertical velocities as well as the adjacency matrix. However, storing an $N \times N$ adjacency matrix is memory-intensive. To bypass this issue, we instead store a matrix of dimension $2 \times E$, where E is the number of edges, compactly encoding the adjacency matrix. This compact representation only stores the two nodes that each edge connects in the graph. This representation is specifically helpful where the adjacency matrix is sparse which is true for the graphs in our dataset, as there are over 1200 nodes in the graph, and each node is only connected to five other nodes on average. 

% This adjacency matrix will be used to encode the spatial relationship between the otherwise uncorrelated velocity node inputs, in the absence of other, more explicit spatial information such as node coordinates or edge lengths. The product of the adjacency matrix with the node level feature matrix creates a single matrix that contains both topological information about the mesh, as well as velocity information.

We perform our approach on two different sets of data. The first study is supposed to only examine the geometry of the airfoils. Therefore, the dataset consists of 1550 different airfoils. The second dataset, however, covers not only different geometries but also various angles of attack. 21 angles of attack are considered for 522 airfoils (10962 samples in total). Angles of attack are changed in the range of \ang{-10} to \ang{10} with an increment of \ang{1}. Given the relatively low velocity in the domain and small angles of attack, the flow regime is laminar, and no significant flow separation occurs.

\begin{figure*}[!tpb]    %[htpb!]
\begin{center}
  \includegraphics[width=1\linewidth]{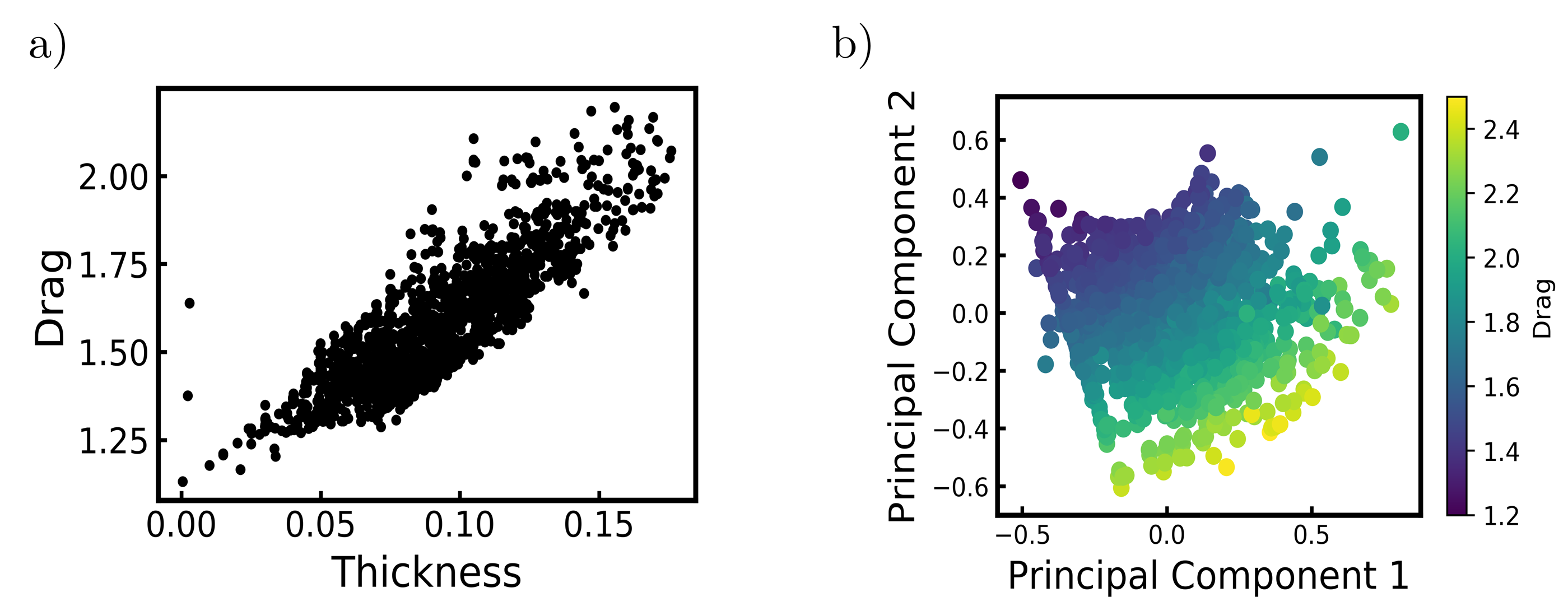}
\end{center}
  \caption{  \textbf{a)} Drag force plotted against the corresponding thickness of 1500 airfoils at a \ang{0} angle of attack. \textbf{b)}
  Principal component analysis on the coordinates of 1500 UIUC airfoils at zero angle of attack, with samples colored by the drag force of the corresponding airfoil.}

\label{fig3}
\end{figure*}

\subsection{Experiments}

Before implementing our method on the airfoil dataset, we perform a study of the relationship between the airfoil geometry and its drag force while other parameters are held constant. Without considering the effect of the angle of attack, we show the drag force to have a positive correlation with the thickness of the airfoil (Figure 3a). However, this correlation, on its own, cannot cover a sufficient portion of the variance in the drag force. In order to determine the geometric features influencing the magnitude of the drag force, we conduct a principal component analysis on the geometry of the airfoil and label the samples by their drag forces. By Principal Component Analysis (PCA), which is a linear dimensionality reduction technique, we extract components that can mostly describe the data variance in an unsupervised manner. While PCA is not generally interpretable, here we can observe the correlation of main components with the drag labels (Figure 3b).

For the first experiment, we use the aforementioned GCNN architecture to predict the drag force for the dataset of airfoils with zero angles of attack. $80\%$ of the samples are randomly selected for training and the remainder are used as a test set. Two complementary metrics of mean squared error (MSE) of drag prediction and the coefficient of determination ($R^2$) are used to quantitatively evaluate the performance. Figure 4a shows the evolution of the loss metric as training progresses. The use of skip-connections in the architecture improves the model's accuracy.

\begin{figure*}[!tpb]    %[htpb!]
\begin{center}
 \includegraphics[width=1\linewidth]{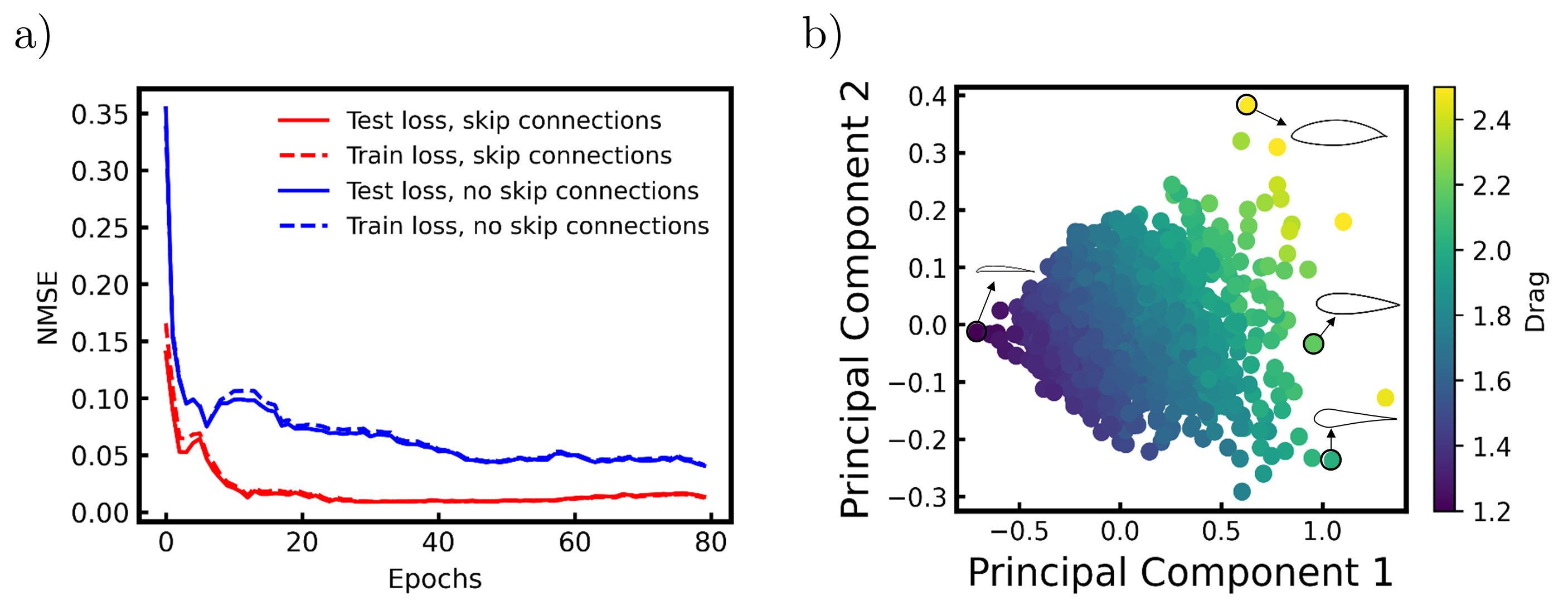}
\end{center}
  \caption{\textbf{a)} The training process of the model. Skip connections increase the accuracy of the model, by reinforcing the information at the output of the final convolutional layer with the embedding created earlier in the model. NMSE: Normalized Mean Square Error. \textbf{b)}   Principal component analysis on the node features of 1500  airfoils at zero angle of attack, after the graph convolution process. Samples are colored by their drag force. Specific airfoils at the extreme of either Principal Component are visualized.}
 \label{fig4}
\end{figure*}

Using node level velocities as the input to the network, the graph convolutional neural network is anticipated to detect the most important features from the flow field data to accurately estimate the drag force. To illustrate
the node embeddings produced by the convolution network, we analyze the values at the input to the first fully connected layer in the trained network which is the averaged output of the convolution layers. 

To do so, we perform a principal component analysis on the features and to detect the two most important geometric components that determine the drag force. It is noteworthy to emphasize that there is no geometrical feature directly encoded in the input of the network. Smooth transition of drag values with two components and depiction of the geometry of samples indicate that the network could learn meaningful geometrical features from flow field data. Here, the first two principal components can explain more than 90\% of the variance in the dataset. The first component encodes a measure of airfoil thickness and the second component is an approximate measure of how quickly the airfoil tapers (Figure 4b).

\begin{figure*}[!tpb]    %[htpb!]
\begin{center}
 \includegraphics[width=0.8\linewidth]{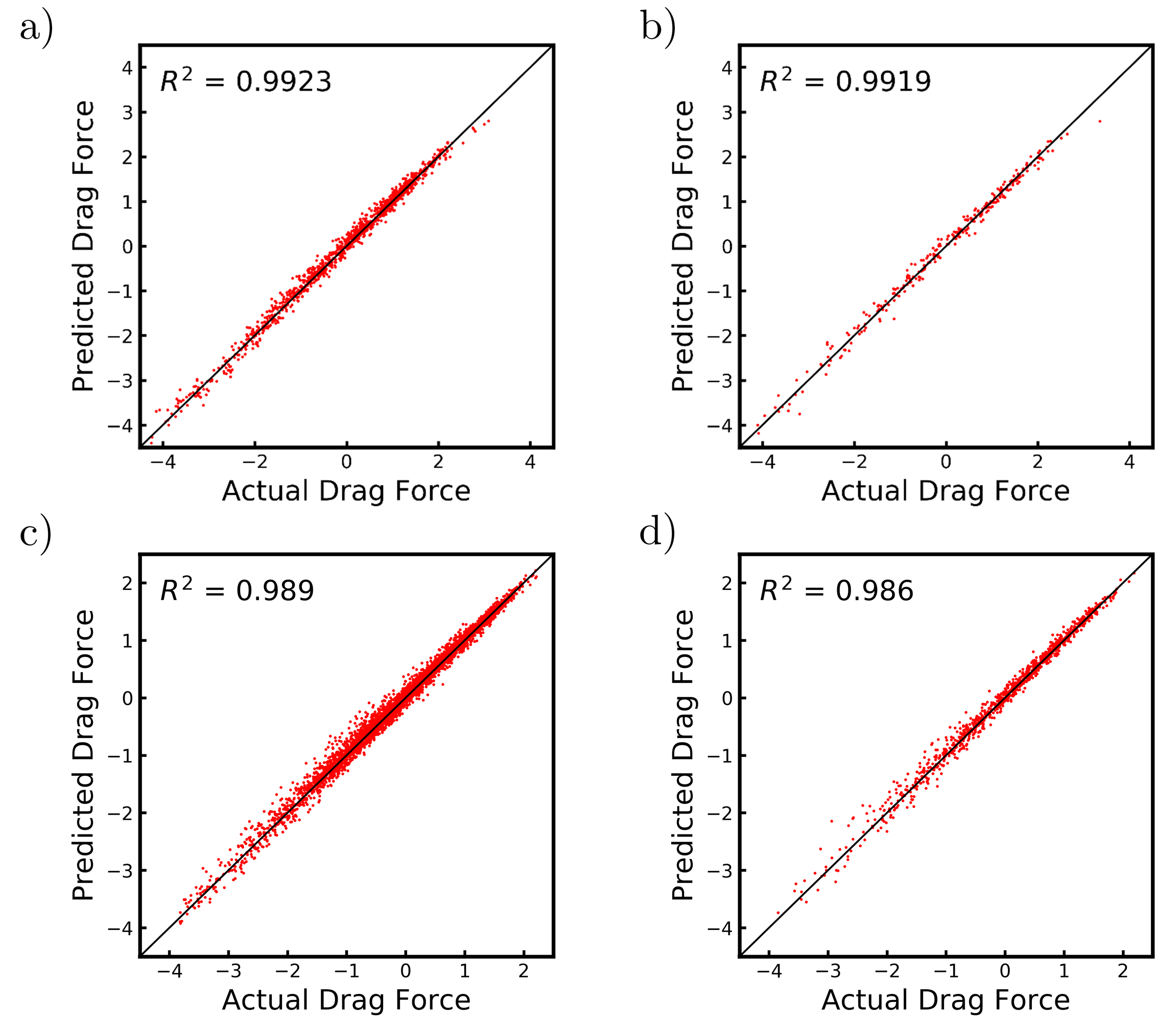}
\end{center}
  \caption{ A comparison of graph convolutional neural network predictions with the ground truth drag force. The first dataset contains 1500 samples of different airfoils at zero angles of attack. \textbf{a)} Data from the training set, \textbf{b)} Data from the test set. The second dataset consists of 5000 samples from 500 airfoils at angles of attack ranging from \ang{-10} to \ang{+10}, with the intervals of \ang{1}. \textbf{c)}: Data from the training set, \textbf{d)}: Data from the test set. }
 \label{fig5}
\end{figure*}
 
\begin{figure*}[!tpb]    %[htpb!]
\begin{center}
 \includegraphics[width=0.5\linewidth]{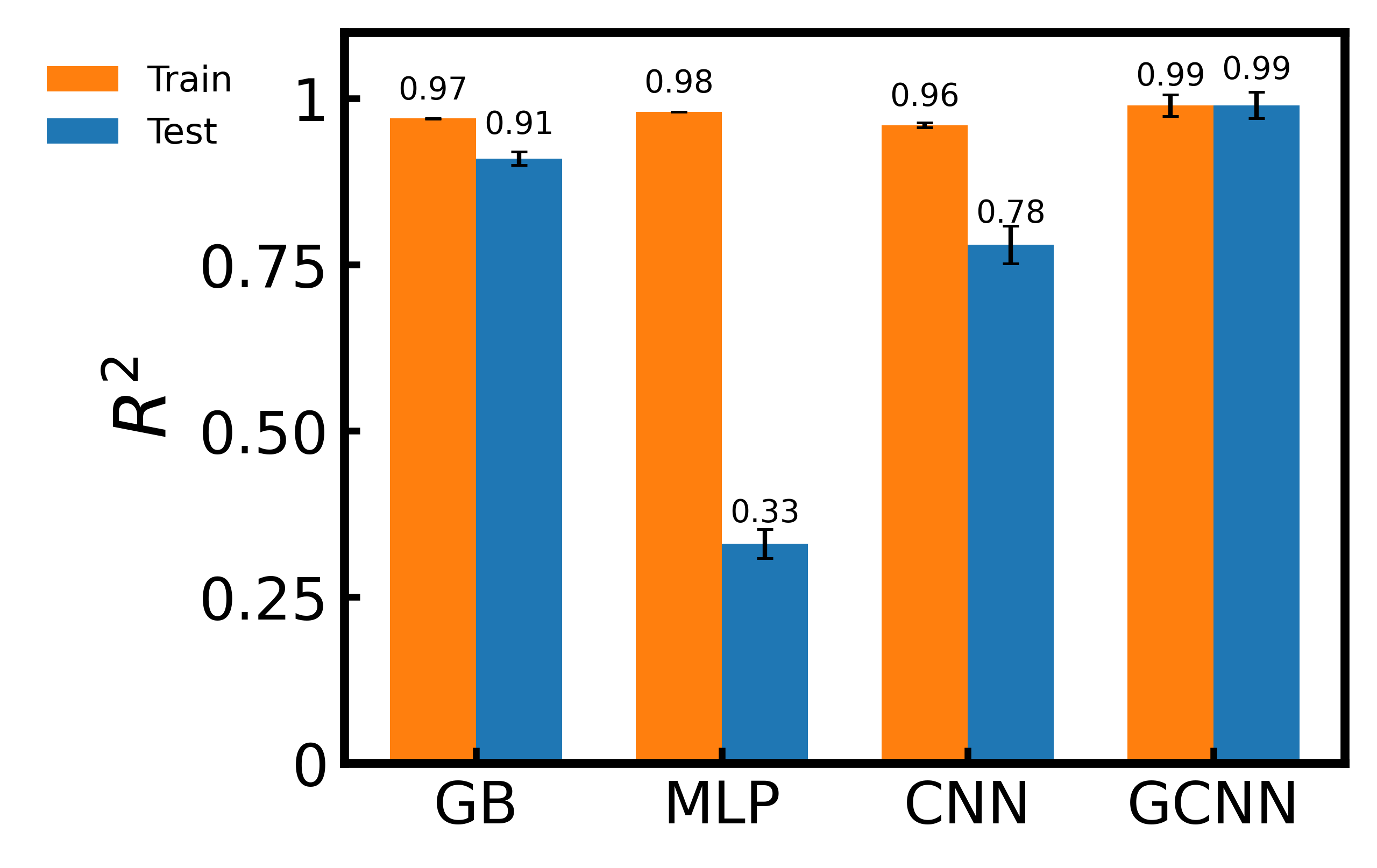}
\end{center}
  \caption{ A comparison of the performance of different prediction algorithms. The dataset consists of 1500  airfoils at zero angle of attack (GB: gradient boosted random forest, MLP: Multilayer perceptron, CNN: Convolutional neural network, GCNN: graph convolutional neural network).}
 \label{fig6}
\end{figure*}

The network is also implemented on the second dataset, containing airfoils that vary in geometry and angle of attack, adding complexity to the prediction task. A comparison of the ground truth values of the drag forces from CFD result and the predictions from the network qualitatively shows the high accuracy of the model for both datasets (Figure 5). 

In addition to GCNNs, other machine learning and neural network algorithms can be applied on the flow field velocity data to solve a regression problem of drag prediction. Notice that the graph size and node order is not a matter of importance for the GCNN as we pass the adjacency and feature matrix directly to the model. However, non-graph based methods require a specified input size, as well as an identical node order between samples. Since the model perceives each input element as a different feature, it cannot be trained unless the order of the node elements are consistent between samples.

\begin{table}[]
\begin{center}
\caption{Details of the trained models}
\renewcommand{\arraystretch}{1.2}
\resizebox{\textwidth}{!}{\begin{tabular}{lll|lll}
\hline
Model              & Parameters             & Settings  & Model          & Parameters & Settings           \\
\hline
{CNN}  & Num Conv layers      & 5       & {MLP} & Num layers     & 4                  \\
  & Conv Kernel      & 3*3       &  &   width layers  & 512                  \\
   & Depth Conv layers      & [64,64,128,256,512]       &  &   Loss   & MSE                  \\ 
& Max Pooling      & 2*2       &  &   Optimiser   & SGD with momentum                 \\ \cline{4-6}
 & Num FC layers      & 3       &  {GB} & Num Estimators &    500      \\ 
 & Width FC layers      & [768,4096,1000,1]       &  & Max Depth &    1     \\ 
\hline
\end{tabular}}
\end{center}
\end{table}

In order to benchmark the GCNN’s performance against traditional, structured machine learning methods, we construct a node ordering  that is consistent across samples. Traditional machine learning methods require the construction of a feature matrix, where the information described by a single feature - i.e., a single column of the feature matrix -  must be consistent from sample to sample. In this formulation, feature i in sample x represents the same information as feature i in sample y.  The node ordering is provided by the mesh generation software based on the order in which the nodes are generated, and is the same for each individual sample. Therefore, since the spatial density of the nodes in each sample is similar, this creates a matrix where Node i in sample x is relatively close in space to Node i as it appears in all other samples. To create a matrix with the same number of features for each sample, the closest 1000 nodes to the center of the airfoil are taken as the feature vector, as there are at least 1000 node measurements in each individual sample. To quantify the spatial similarity of nodes of the same index in this dataset, the distance between nodes of the same index are computed. After comparing the distance from node $i$ in sample $x$ with node $i$ across all other samples, 98\% of these distances are smaller than $5 \times 10^{-3} L$, where $L$ is the length of the domain. This indicates that the position of an arbitrary node is approximately consistent across samples.

To test the performance of non-graph based methods, we select a variety of the most used methods for performing prediction. Therefore, we compare the performance of Gradient Boosted Random Forest regression, a fully connected neural network, and a two-dimensional convolutional neural network on predicting the drag force based on a matrix of node features that adhere to the previously defined structure. Some basic details of these models are provided in Table 3. A  comparison shows that the GCNN approach outperforms the non-graph based methods (Figure 6). 

\section{Conclusion}

We have introduced a novel approach based on graph convolutional neural networks for data-driven prediction using unstructured field information. This method is able to take advantage of the properties of convolution, such as automatic feature detection and parameter sharing while being applied to unstructured data. Flow field properties are usually measured on sparsely scattered points, leading to unstructured data that are incompatible with traditional machine learning algorithms as they only can be applied to structured data. To evaluate the proposed model, the drag force of two-dimensional airfoils are estimated based on the horizontal and vertical components of the flow velocities, measured on the nodes of the irregular mesh around the airfoils. The result of this experiment demonstrates the capability of this approach for global property prediction based on flow field data in similar scenarios. Our model can potentially be extended to experimental cases where access to certain flow information is not readily available. With the currently implemented framework, only velocity information is used to calculate the drag. For instance, the required velocity information can be determined experimentally by analyzing the motion of a sparse set of tracer particles in the flow.  By formalizing the edge connection between tracer observations as a connection from each measurement to the k nearest measurements, one can extend the framework of the Graph Convolutional Neural Network to predict body forces.

The proposed idea of graph representation of the flow field data can be further used for prediction or classification of other field properties whether they are global, such as the drag force, or locally defined on the field. The algorithm can also be used for optimizing the desired properties for design and control applications.

%-------------------------------------------------------------------------

\section*{Funding Sources}
The authors declare that they have no known competing financial interests or personal relationships that could have appeared to influence the work reported in this paper. This work is supported by the start-up fund provided by CMU Mechanical Engineering and funding from Sandia National Laboratories.

\section*{Acknowledgments}
 The authors would like to thank Rishikesh Magar and Yuyang Wang for valuable comments and edits.

\bibliographystyle{refstyle.bst}

\bibliography{sample}

\end{document}